\documentclass[10pt,twocolumn,letterpaper]{article}

\usepackage{cvpr}              

\usepackage[accsupp]{axessibility}

\usepackage{booktabs}       
\usepackage{amsfonts}       
\usepackage{nicefrac}       
\usepackage{microtype}      
\usepackage{xcolor}         
\usepackage{multirow}
\usepackage{soul} 
\usepackage{color}
\definecolor{Gray}{gray}{0.9}
\definecolor{Gray2}{gray}{0.95}
\usepackage{color,colortbl}
\usepackage{subcaption}
\usepackage[pdftex]{graphicx}
\usepackage{pifont}
\usepackage{amsmath}
\usepackage{amssymb}
\usepackage{rotating}
\newcommand{\spheading}[2][6.4em]{
  \rotatebox{90}{\parbox{#1}{\raggedright #2}}}

\usepackage[pagebackref,breaklinks,colorlinks]{hyperref}

\usepackage[capitalize]{cleveref}
\crefname{section}{Sec.}{Secs.}
\Crefname{section}{Section}{Sections}
\Crefname{table}{Table}{Tables}
\crefname{table}{Tab.}{Tabs.}

\def\cvprPaperID{10013} 
\def\confName{CVPR}
\def\confYear{2023}

\begin{document}

\title{MammalNet: A Large-scale Video Benchmark \\for Mammal Recognition and Behavior Understanding} 

\author{Jun Chen$^1$$^*$ \qquad Ming Hu$^1$$^*$ \qquad Darren J. Coker$^1$ \qquad
Michael L. Berumen$^1$  \\ Blair Costelloe$^{2,3}$ \qquad Sara Beery$^4$ \qquad Anna Rohrbach$^5$ \qquad  Mohamed Elhoseiny$^1$\\ \textsuperscript{\rm 1}{King Abdullah University of Science and Technology (KAUST)} \\
   \textsuperscript{\rm 2}
  {Max Planck Institute of Animal Behavior },  \,\, \textsuperscript{\rm 3}{ University of Konstanz} \\  
\textsuperscript{\rm 4}{Massachusetts Institute of Technology}, \,\, \textsuperscript{\rm 5}{University of California, Berkeley} 
}

\makeatletter
\let\@oldmaketitle\@maketitle
\renewcommand{\@maketitle}{\@oldmaketitle
\vspace{-5pt}
\centering
\includegraphics[width=0.8\linewidth]{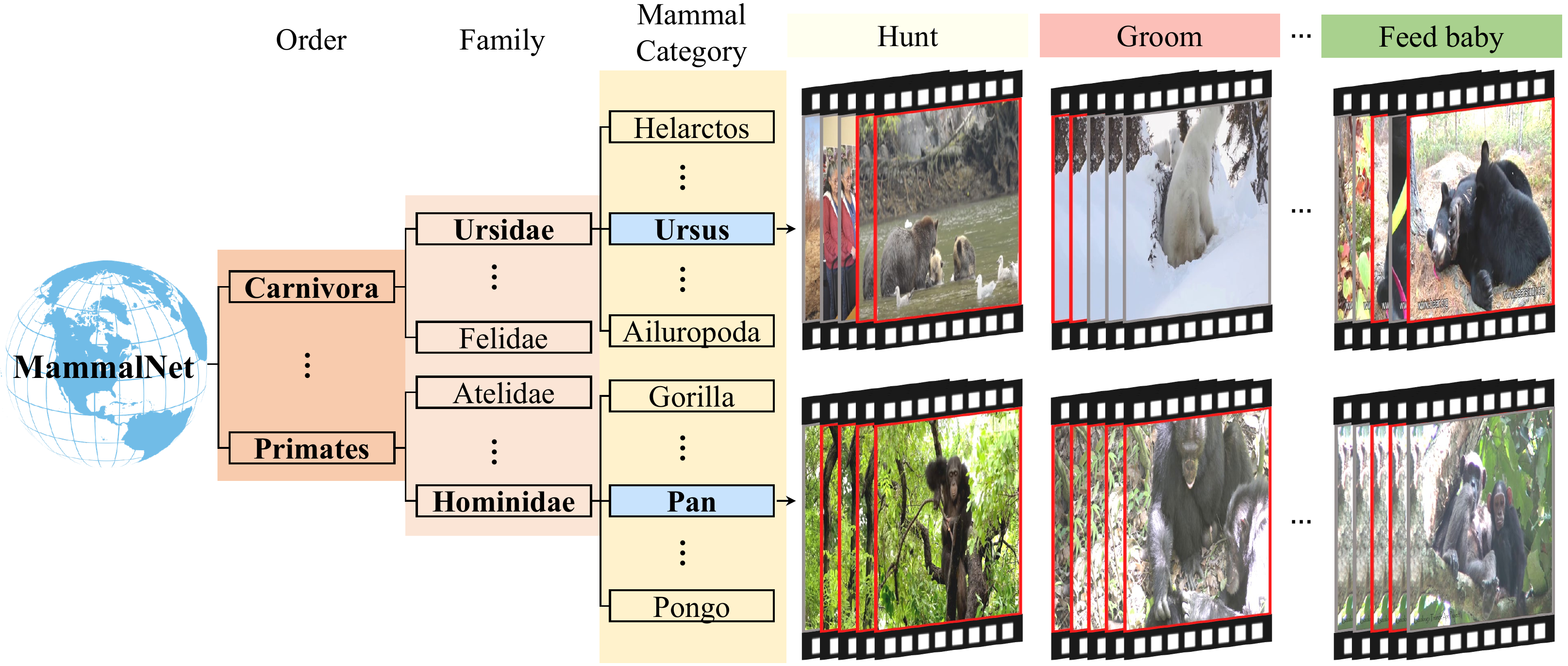}
\vspace{-2mm}
\captionof{figure}{We propose MammalNet, a large-scale video benchmark for recognizing mammals and their behavior. It is built around a biological mammal taxonomy spanning 17 orders, 69 families and 173 mammal categories, and includes 12 common high-level mammal behaviors (e.g., hunt, groom). MammalNet enables the study of animal and behavior recognition, both separately and jointly. It also facilitates investigating challenging compositional scenarios which test models' zero- and low-shot transfer abilities. Moreover, MammalNet includes behavior detection by localizing when a behavior occurs in an untrimmed video. Our dataset is the first to enable animal behavior analysis at scale in an ecologically-grounded manner, and exemplifies multiple challenges for the computer vision community, such as recognition of imbalanced, hierarchical distributions of fine-grained categories and generalization to unseen or seldom seen scenarios.
}
\vspace{-4mm}
\label{fig:teaser}
\bigskip\bigskip}
\makeatother

\maketitle

%

\let\thefootnote\relax\footnotetext{ $^*$Equal contribution.}
\begin{abstract}

Monitoring animal behavior can facilitate conservation efforts by providing key insights into wildlife health, population status, and ecosystem function.  Automatic recognition of animals and their behaviors is critical for capitalizing on the large unlabeled datasets generated by modern video devices and for accelerating monitoring efforts at scale. However, the development of automated recognition systems is currently hindered by a lack of appropriately labeled datasets.
Existing video datasets 1) do not classify animals according to established biological taxonomies; 2) are too small to facilitate large-scale behavioral studies and are often limited to a single species; and 3) do not feature temporally localized annotations and therefore do not facilitate localization of targeted behaviors within longer video sequences.
Thus, we propose \emph{MammalNet}, a new large-scale animal behavior dataset with taxonomy-guided annotations of mammals and their common behaviors. MammalNet contains over 18K videos totaling 539 hours, which is $\sim$10 times larger than the largest existing animal behavior dataset \cite{ng2022animal}. It covers 17 orders, 69 families, and 173 mammal categories for animal categorization and captures 12 high-level animal behaviors that received focus in previous animal behavior studies. We establish three benchmarks on MammalNet: standard animal and behavior recognition, compositional low-shot animal and behavior recognition, and behavior detection. Our dataset and code have been made available at: \url{https://mammal-net.github.io}.

\end{abstract}

\section{Introduction}


\begin{table*}[t!]
\small
\begin{center}
\resizebox{0.9\linewidth}{!}{
\begin{tabular}
{p{0.16\linewidth}|p{0.03\linewidth}|p{0.06\linewidth}|p{0.05\linewidth}|p{0.04\linewidth}|p{0.05\linewidth}| 
p{0.05\linewidth}|p{0.06\linewidth}|p{0.06\linewidth}|p{0.06\linewidth}|p{0.06\linewidth}|p{0.06\linewidth}}
\toprule
\multirow{8}{*}{Datasets} &\multicolumn{8}{c|}{Dataset Properties} & \multicolumn{3}{c}{Tasks} \\  
&\cline{1-11}
& \spheading{Publicly Available?} & \spheading{Taxonomy-guided Animal Annotation?} &\spheading{ No. of Videos} & \spheading{No. of Actions} & \spheading{No. of Behaviors} 
& \spheading{No. of Animal Categories}
& \spheading{No. of Mammal Categories}  &\spheading{Total Duration } 
& \spheading{Animal Classification} 
& \spheading{Action/Behavior Recognition} & \spheading{Action/Behavior  Detection} \\
\midrule
Wild Felines \cite{feng2021action} &\ding{53} &\ding{53} & 2,700 & 3  & -
& 3 
&3 & -&$\checkmark$ &  $\checkmark$  & \ding{53} \\
Wildlife Actions \cite{li2020wildlife} & \ding{53}& \ding{53}  & 10,600 & 7 & - & 
32&
11& - & $\checkmark$& $\checkmark$ & \ding{53}\\ 
  Animal Kingdom \cite{ng2022animal} & $\checkmark$ & \ding{53}& 4,301 & 140 & -
  & 850
  & 
  -
  &  50 (h) & \ding{53} & $\checkmark$& \ding{53}\\
  MammalNet (ours) & $\checkmark$ & $\checkmark$ & 18,346 & -& 12
  & 173 
  & 173 & 539 (h)& $\checkmark$& $\checkmark$ & $\checkmark$ \\
\bottomrule
\end{tabular}}
\end{center}
\caption{The comparison among existing animal behavior understanding video datasets. Compared to other datasets, MammalNet annotates the animals by following the scientific mammal taxonomy, focuses on more high-level behavior recognition,  has the largest number of mammal categories, collects the largest number of animal behavior videos, totalling 539 hours, and also enables behavior detection tasks.} 
\label{dataset_comparison}
\vspace{-1.5em}
\end{table*}

Animal species are a core component of the world's ecosystems. Through their behavior, animals drive diverse ecological processes, including seed dispersal, nutrient cycling, population dynamics, speciation, and extinction. Thus, understanding and monitoring the behaviors of animals and their interactions with their physical and social environments is key to understanding the complexities of the world's ecosystems, an objective that is especially critical now given the ongoing biodiversity crisis~\cite{ceballos2020vertebrates}.

Modern sensors, including camera traps, drones, and smartphones, allow wildlife researchers, managers, and citizen scientists to collect video data of animal behavior on an unprecedented scale~\cite{tuia2022perspectives}. However, processing this data to generate actionable, timely insights remains a major challenge. Manual human review and annotation of footage to identify and locate species and behavioral sequences of interest is time-intensive and does not scale to large datasets. Thus, methods for automated animal and behavioral recognition could open the door to large-scale behavioral monitoring and speed up the time to produce usable data, thereby reducing the time to implement management directives.

The first essential step to creating such an AI system for animal and behavior recognition is curating a diverse, representative dataset that allows us to formalize these challenges as computer vision tasks and benchmark potential solutions. Most previous datasets either only cover a limited number of animal and behavior types~\cite{anderson2014toward,rahman2014fast}, or do not implement animal labeling~\cite{ng2022animal}, or include a small number of videos with insufficient environmental diversity~\cite{von2021big,anderson2014toward,rahman2014fast}. Recently, a dataset named ``Animal Kingdom''~\cite{ng2022animal} was proposed to study animal actions and is currently the largest existing behavioral dataset, to the best of our knowledge. However, it only contains 4,310 videos totaling 50 hours, which might be insufficient for large-scale animal behavior studies considering its diversity. Furthermore, the authors only focus on the recognition of atomic actions such as yawning, swimming, and flying. These basic actions cannot be easily matched to the higher-order behavioral states that are of primary interest to end users in animal management and conservation \cite{animal_behavior}. For example, a cheetah that is running may either be hunting, escaping, or playing. Finally, and most importantly, they do not 
 support some important tasks such as animal recognition and behavior detection which are essential for animal behavior understanding. 


To overcome the limitations of previous datasets, we propose a new dataset called \emph{MammalNet}. We specifically focus on mammals since they, unlike other animal classes such as birds or insects, usually have more diverse and distinguishable behavior statuses. MammalNet is comprised of 539 hours of annotated videos, which is $\sim$10 times longer than that of the largest available animal behavior dataset. It contains 18,346 videos depicting 12 fundamental high-level behaviors from hundreds of mammal species. Importantly, 
it focuses on 12 higher-order animal behaviors that are the focus of previous animal behavior literature~\cite{breed2021animal,dugatkin2020principles,alcock2009animal,mench1998important}, rather than atomic actions.
MammalNet also categorizes animals according to the scientific taxonomy available in Wikipedia, as we show in Fig.~\ref{fig:teaser}; hence the dataset can be flexibly expanded in the future by following the same protocols. 
It includes videos of approximately 800 mammal species in 173 mammal categories.
We establish three benchmarks inspired by ecological research needs - standard animal \& behavior classification, compositional low-shot animal \& behavior recognition, and behavior detection – to promote future study in animal  behavior understanding. 

Through our experiments, we find that: (1) Correctly recognizing the animals and behaviors is a challenging task even for the state-of-the-art models, especially for less-frequent animals. The top-1 per-class accuracy is 32.5 for animal recognition, 37.8 for behavior recognition, and 17.8 for their joint recognition in our best-performing model. (2) Behavior recognition for unseen animals can be transferred from observations of other seen animals due to their similar features such as appearance and movement style, which can help in studies of animals with less available data. However, to achieve more accurate behavior recognition, having access to videos of the target animals and behaviors is still crucial.


\section{Related Work}

Automatic animal recognition and behavior detection can help humans monitor and efficiently process animal behavior data~\cite{mathis2020deep,robinson2014comparison,von2021big,gupta2021dftnet,mu2020learning,beyan2013detection,van2022ssw60}. It can massively reduce the labour cost from manually collecting and analyzing  animal activities. During the past few years, many datasets~\cite{rahman2014fast,feng2021action,li2020wildlife,fang2021pose} have been introduced to develop foundations for animal behavior research. We systematically analyze the previous datasets and summarize several important limitations that prevent them from being used for large-scale animal recognition and behavior understanding:

\textbf{Lack of behavior understanding.} Many previous datasets only focus on animal recognition or pose estimation from images, but lack behavior learning. For example, iNaturalist~\cite{inaturalist} collects 859,000 images covering more than 5,000 different types of plants and animals. NABird~\cite{van2015building} collects 48,562 images of 555 different North American bird species. Also, some works focus on narrow animal recognition such as dogs~\cite{khosla2011novel}, birds~\cite{welinder2010caltech,vanexploring} or cats~\cite{chen2016locality}. On the other hand, many works also focus on pose estimation~\cite{mathis2021pretraining,shooter2021sydog,graving2019deepposekit,cao2019cross,fang2021pose} and animal face detection~\cite{khan2020animalweb,yang2016human,rashid2017interspecies}. They are generally not applicable to learning animal behavior.

\textbf{Lack of taxonomic diversity for different behaviors.} Previous animal behavior datasets have minimal taxonomic coverage - often containing just a single animal species. For example, there are existing behavior recognition datasets for elephants~\cite{laws2007case}, sheep~\cite{owoeye2018online}, monkeys~\cite{bala2020automated}, tigers~\cite{dishman2014self}, etc. While useful for the studied species, these datasets do not enable the exploration of behavior across species which is necessary to scale up behavior identification without requiring training examples of every possible combination of species and behavior.

\textbf{Lack of taxonomy-guided animal annotation.} Previous animal behavior datasets either do not include animal recognition as a task~\cite{ng2022animal} or group species according to subjective as opposed to scientific criteria~\cite{li2020wildlife,ng2022animal}. In contrast, our dataset collects and annotates the videos according to the scientific mammal taxonomy. By following this taxonomy we allow exploration of behavior from an evolutionary perspective, as well as enable standardized and consistent dataset expansion under the same protocol.

\textbf{Actions vs. Behaviors.} Previous works mostly focus on atomic action recognition~\cite{ng2022animal, li2020wildlife,feng2021action}. For example, some of their action classes are \emph{walk, stand still, fly, etc}. In contrast, MammalNet focuses on higher-level animal behaviors, such as \emph{hunt, feed baby, etc}. Behavior here denotes the main activity instance during a period, and it is usually composed of multiple atomic actions. These complex behaviors are needed to describe and summarize video activity bouts in a manner which is valuable for ecological research. 

We further compare our MammalNet dataset with the other existing animal behavior understanding video datasets, and summarize the key difference in Table \ref{dataset_comparison}.


\section{Constructing MammalNet}

\begin{figure}[t!]
\centering
\includegraphics[width=0.9\linewidth]
                  {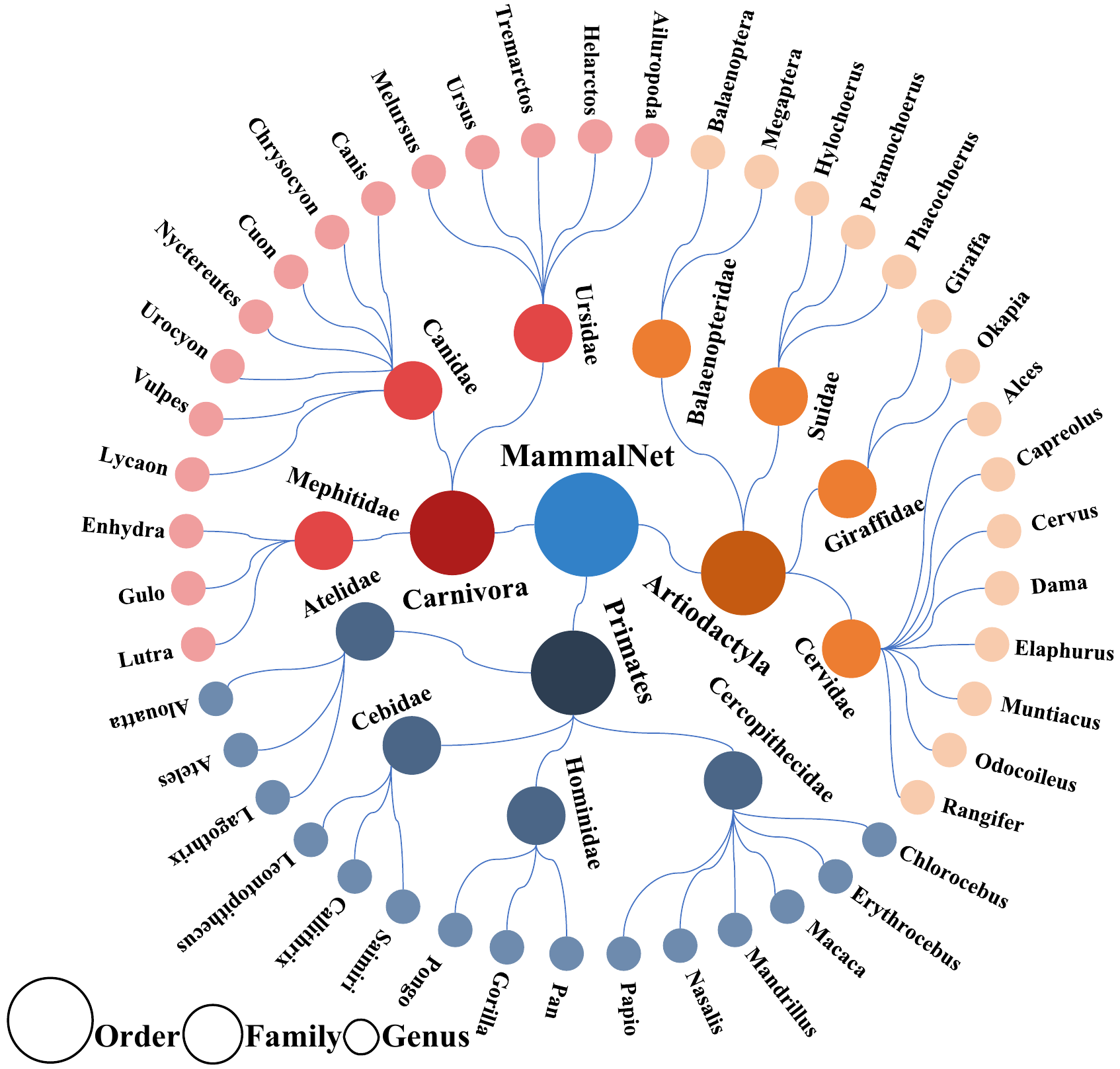}
\caption{A subset of the mammal taxonomy of MammalNet. It includes 3 orders, 11 families, and 45 genera.}
\label{animal_taxonomy}
\vspace{-4mm}
\end{figure}

The goal of \emph{MammalNet} is to provide a large-scale mammal video dataset that benchmarks both animal and behavior recognition. In this section, we discuss our dataset construction protocol, including the choice of scientific animal taxonomy, crowdsourced annotations, and performing manual quality control during video collection and annotation. Finally, we describe the statistical profile of MammalNet.

\begin{figure*}[t!]
\centering
\includegraphics[width=0.85\linewidth]
                  {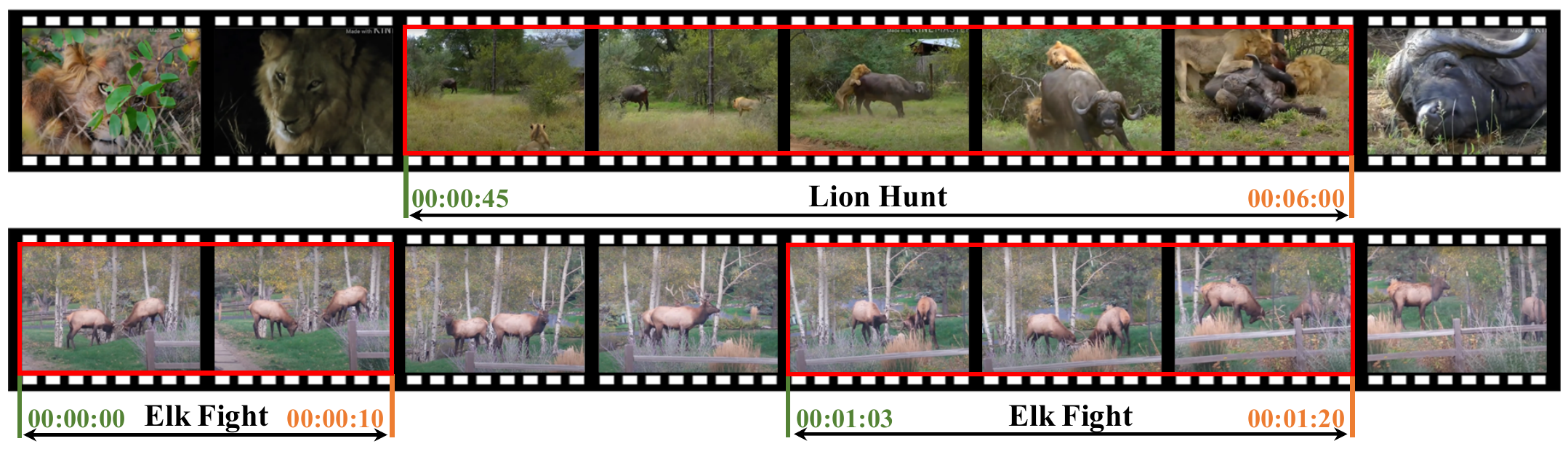}
\caption{The examples for the annotated target behavior boundaries. The frames marked in red boxes denote the annotated temporal boundaries for the target behavior.}
\label{behavior_boundary}
\vspace{-4mm}
\end{figure*}

\subsection{Animal Taxonomy and Behavior Collection} 
\textbf{Animal taxonomy construction.} Accurate animal recognition is one of the main goals defined by MammalNet. To ensure this task is scientifically relevant, we aimed to collect and structure our list of target animals based on the mammal taxonomy. 
First, we collected a diverse list of mammals, approximately 800 mammal species and sub-species, from the \textbf{National Geographic}~\cite{national_geographic} and \textbf{Animal A-Z} \cite{animal_a_z} websites. Next, we manually mapped each mammal onto the taxonomic structure (including class, order, family, genus, tribe, sub-family and species) in mammal taxonomy from Wikipedia. In total, the classes included in MammalNet cover 17 orders, 69 families, and 173 mammal categories. A taxonomic subset is visualized in Fig.~\ref{animal_taxonomy}.

\textbf{Behavior collection.} We aim to enable the study of complex, high-level animal behavior as opposed to the simpler atomic actions emphasized in previous work~\cite{ng2022animal,li2020wildlife}. Behavior here represents the major activity being displayed during a period of a video, and can be viewed as a series of atomic actions collectively serving a higher-level purpose. For example, hunting behavior, as defined in MammalNet, can often be decomposed into running, chasing and killing actions, etc. Identification of behaviors at this level of definition is more useful for ecologists and zoologists~\cite{breed2021animal,dugatkin2020principles,alcock2009animal,mench1998important} compared to atomic actions. 

Inspired by previous biological and ecological studies~\cite{breed2021animal,dugatkin2020principles,alcock2009animal}, we consider 12 fundamental mammal behaviors under 5 different groups in our study. They are respectively: 

\noindent \textbf{Foraging behaviors}: eat food, drink water, hunt. 

\noindent \textbf{Reproductive behaviors}: mate, feed baby, give birth. 

\noindent \textbf{Hygiene behaviors}: groom.

\noindent \textbf{Agonistic behaviours}: fight. 

\noindent \textbf{Maintenance behaviors}: urinate, defecate, sleep, vomit.

\begin{figure*}[t]
\centering
\includegraphics[width=0.85\linewidth]
                  {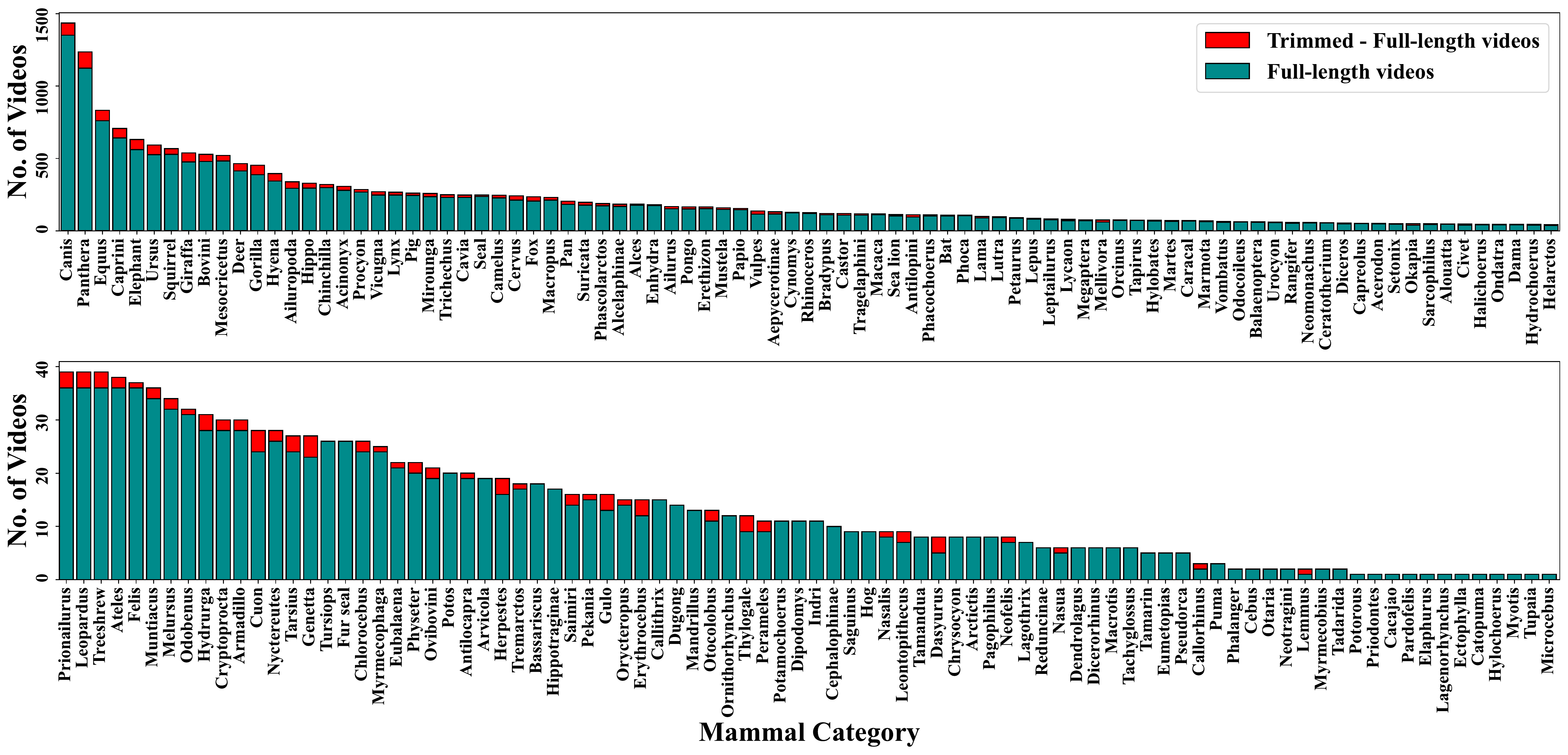}
\caption{Number of videos per each mammal category. We rank the categories according to their trimmed videos frequency.}
\label{per_genus}
\vspace{-4mm}
\end{figure*}
\begin{figure}[t]
\centering
\includegraphics[width=0.7\linewidth]
                  {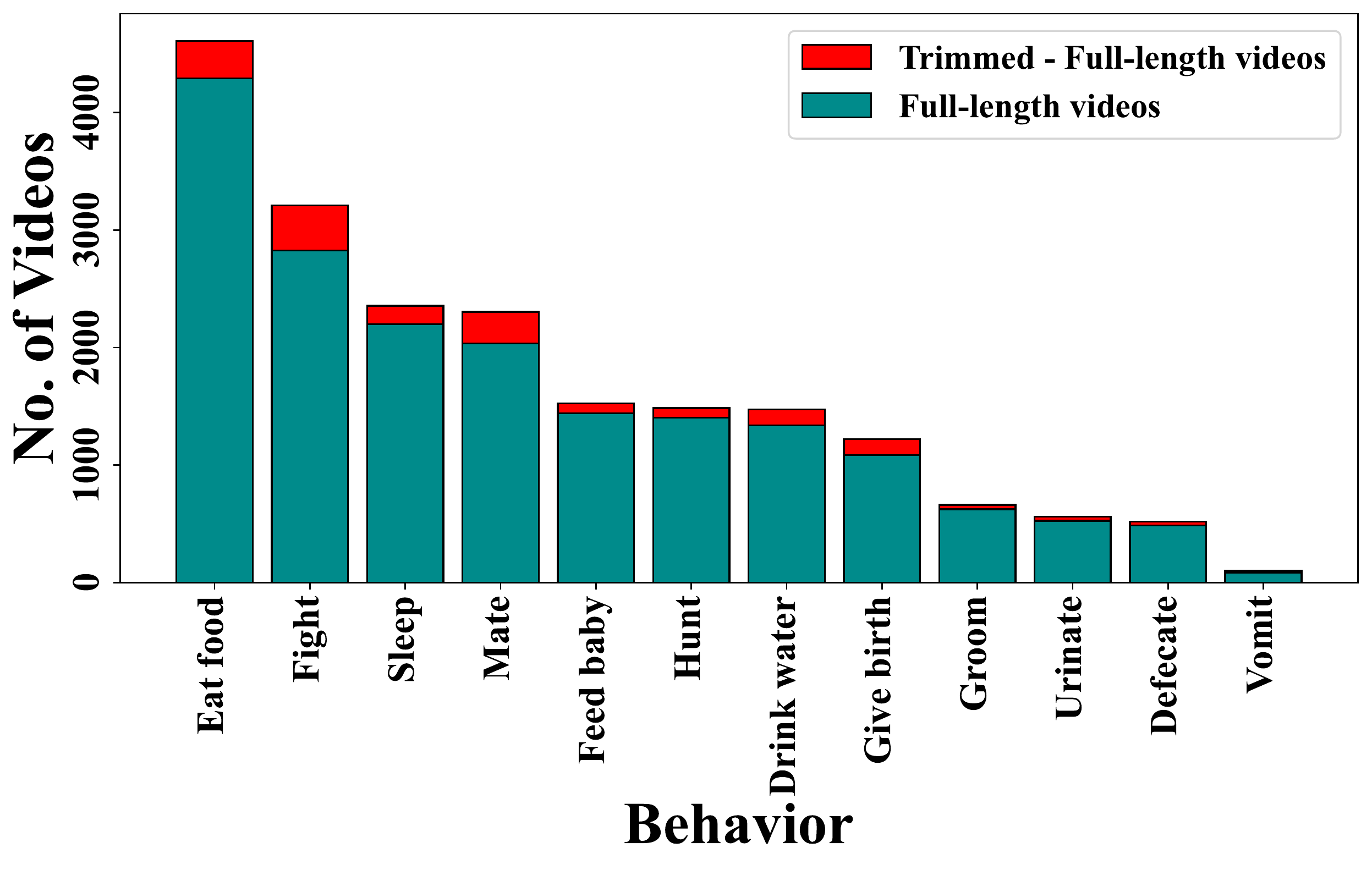}
\caption{Number of videos per each behavior. We rank the behavior according to their trimmed videos frequency.}
\label{per_behavior}
\vspace{-4mm}
\end{figure}

\subsection{Video Collection and Quality Assurance}
Dataset curation is usually a costly process requiring a lot of manual annotation by humans. Some datasets are annotated by domain experts to encourage more reliable labeling, particularly for challenging tasks, but this expertise comes at a much higher cost and is thus hard to scale. We adapted a semi-automatic crowdsourcing approach to collect and annotate our datasets, inspired by many previous works~\cite{deng2009imagenet,heilbron2014collecting,vondrick2013efficiently,caba2015activitynet}.

\textbf{Online video retrieval.} Our goal was to collect videos depicting each animal in our database performing each of the 12 focal behaviors. To achieve this, we queried YouTube with the text combinations of each animal common name and behavior, e.g., \emph{tiger hunting}, and downloaded videos where the title included our queried animal name and behavior. However, some animal names are too rare to generate enough relevant videos, e.g., \emph{Brown hyena} or \emph{Spotted hyena}. In such cases, we used a more common name, \emph{hyena}, to represent the union of those animals. In order to retrieve more relevant videos from the search engine, we also expanded each animal with their synonyms as given in Wikipedia. For example, we expanded the original \emph{artic fox} with other equivalent names such as \emph{white fox}, \emph{polar fox} and \emph{snow fox}. Each video was downloaded at its highest available resolution.

\textbf{Data filtering.} During video retrieval, we downloaded the videos that are accessible for people $<$16 years old to avoid the videos with violent content. We also prioritized videos that are shorter than 10 minutes in duration to limit the total storage.

However, some videos might have irrelevant content due to the inaccuracy of text-based retrieval. For example, the downloaded videos might 1) display cartoon or toy animals or unrealistic environments such as games or movies, 2) display a static image instead of a continuous video, 3) involve a lot of human-animal interaction, e.g., human feeding an animal, as opposed to focusing on animal behavior, 4) not contain the specified animal and/or behavior.

To alleviate these issues, we employed Amazon Mechanical Turk workers to verify the presence of the animal and behavior, and identify other quality issues. We assigned each video to three different workers and provided them with the pictures of an animal with its common name, and an expected behavior. We asked the workers to verify if this animal and behavior indeed appear in the video. Only the videos that ``pass'' by all the three workers were kept for the following behavior localization annotation. Before the workers started the verification, we first provided them the verification instructions and asked them to complete a corresponding qualification test with 20 multiple choice questions to ensure only qualified workers could participate in our task.

\subsection{Animal Behavior Localization Annotation} The target behavior typically does not span the whole video. To localize the part where the target behavior is actually occurring in the video, we asked the AMT workers to manually annotate the respective temporal boundaries. Namely, we asked five different qualified workers to annotate the start and end frames for the target behavior in each video. In the end, each video received at least 5 annotated behavior boundaries. To achieve robust annotation agreement, we used the complete linkage algorithm~\cite{defays1977efficient} to cluster different temporal boundaries and merge them into one or several more stable ones that received multiple agreements. 
Note, that a single video might have multiple discrete occurrences of a given behavior, and thus have multiple boundary definitions.  We show several examples for annotated target behavior boundaries in Fig.~\ref{behavior_boundary}.

\subsection{Recognition at lowest feasible taxonomic level}
We categorize animals according to the lowest and feasible taxonomic classification, rather than the species level, for the following reasons: 1) YouTube videos contain the ambiguous and inexpert species labels. 2) We are not using expert annotators, and even experts often cannot reliably identify animals at the species or even genus level based on crowdsourced or field images~\cite{kays2022mammals}. 
 Thus, it is more practical to classify animals in our dataset to the lowest feasible taxonomic level rather than species. As a result, our taxonomy contains 173 distinguishable taxonomic classification levels including:  sub-family, tribe and genus.

\subsection{MammalNet Statistics}

The final MammalNet dataset contains 18,346 videos (539 hours); after we trim the videos according to the annotated behavior boundaries, it increases to 20,033 trimmed video instances (394 hours). In total, it covers 173 mammal categories, 69 families and 17 orders in our dataset, with the total of 173 mammal categories defined for our recognition tasks. MammalNet also contains 12 different common behaviors. The average duration for the untrimmed and trimmed videos is 106 and 77 seconds, respectively. Over 54\% of videos reach HD resolutions (1280 $\times$ 720). We demonstrate the data distribution in terms of each animal category and behavior in Fig.~\ref{per_genus} and ~\ref{per_behavior}. We observe that the number of collected videos per type follow a long-tail distribution.





\begin{table*}[t!]
\small 
\begin{center}
\begin{tabular}{lcccccccccccc}
\toprule
&\multicolumn{4}{c}{Animal Classification} 
&\multicolumn{4}{c}{Behavior Classification}
&\multicolumn{4}{c}{Joint Classification}\\
\multirow{1}{*}{Baselines} & \multicolumn{1}{c}{Many} & \multicolumn{1}{c}{Medium} & \multicolumn{1}{c}{Few} &
\multicolumn{1}{c}{All} &
\multicolumn{1}{c}{Many} & \multicolumn{1}{c}{Medium} & \multicolumn{1}{c}{Few} &
\multicolumn{1}{c}{All} &
\multicolumn{1}{c}{Many} & \multicolumn{1}{c}{Medium} & \multicolumn{1}{c}{Few} &
\multicolumn{1}{c}{All}\\
 & 12  & 28  & 133 &  173 & 4 & 4 & 4 & 12 & 33 & 180 & 823 & 1036\\
\midrule
SlowFast \cite{feichtenhofer2019slowfast} & \textbf{49.6} & \textbf{35.6} & 9.5 & 17.2 & 39.0 & 27.6 & \textbf{14.9} &27.2 & 27.2 & 19.4 & 4.4 & 8.6 \\
C3D \cite{c3d} & 48.6 & 35.3 & 10.0 & 17.5 & 38.2 & 27.8 & 11.7 & 25.9 & 28.1 & 18.8 & 4.5 & 8.6 \\
I3D \cite{i3d} & 48.8  & 34.9 & 10.5 & 17.8 & 39.5 & 27.2 & 14.8 & 27.2  & \textbf{29.6} & \textbf{20.5} & 4.4 & 8.9 \\
MViT V2 \cite{li2022mvitv2} & 48.5 & 35.5 & \textbf{12.9} & \textbf{19.7} & \textbf{42.3} & \textbf{29.2} & 11.6 & \textbf{27.7} &29.5 & 19.6 & \textbf{4.8} & \textbf{9.0}\\

\midrule
SlowFast*  & 58.3  & 43.1 & 16.6 & 24.5 & 45.1 & 32.7 & 14.8 & 30.9 & 38.6 & 23.5 & 7.3 & 12.1 \\
C3D*  & 58.3 & 45.4 & 19.1 & 26.8 & 44.6 & 36.0 & 15.9 & 32.2 & 38.0 & 26.4 & 8.4 & 13.5  \\
I3D*  & 58.6  & 42.9 & 16.9 &24.8 &46.3  & 35.0 & 14.8 & 32.1   & 38.3 & 24.5 & 8.6 & 13.3\\
MViT V2* & \textbf{66.7} & \textbf{56.0} & \textbf{23.4} & \textbf{32.5}  & \textbf{50.9} & \textbf{42.4}  & \textbf{20.0}& \textbf{37.8}  & \textbf{46.2} &  \textbf{33.0} & \textbf{11.8} & \textbf{17.8}\\
\bottomrule
\end{tabular}
\end{center}
\caption{Per-class Top-1 accuracy for animal, behavior and their joint prediction.* denotes the initialization from the model pretrained on Kinetics 400~\cite{kay2017kinetics}. Transfer learning from the human action to the animal behavior recognition receives considerable performance gain. Best performance for each split has been highlighted in \textbf{bold}. } 
\label{per_class_accuracy}
\vspace{-4mm}
\end{table*}

\begin{table*}[t!]
\small 
\begin{center}
\begin{tabular}{lcccccccccccc}
\toprule
&\multicolumn{4}{c}{Animal Classification} 
&\multicolumn{4}{c}{Behavior Classification}
&\multicolumn{4}{c}{Joint Classification}\\
\multirow{1}{*}{Baselines} & \multicolumn{1}{c}{Many} & \multicolumn{1}{c}{Medium} & \multicolumn{1}{c}{Few} &
\multicolumn{1}{c}{All} &
\multicolumn{1}{c}{Many} & \multicolumn{1}{c}{Medium} & \multicolumn{1}{c}{Few} &
\multicolumn{1}{c}{All} &
\multicolumn{1}{c}{Many} & \multicolumn{1}{c}{Medium} & \multicolumn{1}{c}{Few} &
\multicolumn{1}{c}{All}\\
 & 12  & 28  & 133 &  173 & 4 & 4 & 4 & 12 & 33 & 180 & 823 & 1036\\
\midrule
SlowFast  & 58.3  & 43.1 & 16.6 & 24.5 & 45.1 & 32.7 & 14.8 & 30.9 & 38.6 & 23.5 & 7.3 & 12.1 \\
C3D  & 58.3 & 45.4 & 19.1 & 26.8 & 44.6 & 36.0 & 15.9 & 32.2 & 38.0 & 26.4 & 8.4 & 13.5  \\
I3D  & 58.6  & 42.9 & 16.9 &24.8 &46.3  & 35.0 & 14.8 & 32.1   & 38.3 & 24.5 & 8.6 & 13.3\\
MViT V2 & \textbf{66.7} & \textbf{56.0} & \textbf{23.4} & \textbf{32.5}  & \textbf{50.9} & \textbf{42.4}  & \textbf{20.0}& \textbf{37.8}  & \textbf{46.2} &  \textbf{33.0} & \textbf{11.8} & \textbf{17.8}\\
\bottomrule
\end{tabular}
\end{center}
\caption{Per-class Top-1 accuracy for animal, behavior and their joint prediction.* denotes the initialization from the model pretrained on Kinetics 400~\cite{kay2017kinetics}. Transfer learning from the human action to the animal behavior recognition receives considerable performance gain. Best performance for each split has been highlighted in \textbf{bold}. } 
\label{per_class_accuracy}
\vspace{-4mm}
\end{table*}

\begin{table}[t]
\small 
\begin{center}
\begin{tabular}{lccc}
\toprule
\multirow{1}{*}{Baselines} &\multicolumn{1}{c}{Animal } 
&\multicolumn{1}{c}{Behavior}
&\multicolumn{1}{c}{Joint }\\
\midrule
SlowFast \cite{feichtenhofer2019slowfast} &  35.4 & 34.2 &  17.4 \\
C3D \cite{c3d} & 35.0 & 33.5 & 17.1 \\
I3D \cite{i3d} & 35.2  & 34.3  & 17.9 \\
MViT V2 \cite{li2022mvitv2} &  \textbf{35.6}  & \textbf{36.8} & \textbf{18.0} \\
\midrule
SlowFast* & 43.0 & 39.4 & 22.8 \\
C3D* & 44.4 & 40.3 & 24.6  \\
I3D* & 43.4 & 41.2 & 24.0 \\
MViT V2* & \textbf{52.6} & \textbf{46.6} & \textbf{30.6} \\
\bottomrule
\end{tabular}
\end{center}
\caption{Per-example accuracy for animal, behavior and their joint prediction. * denotes the initialization from the pretrained model.} 
\label{per-example-accuracy}
\vspace{-2em}
\end{table}

\section{Experimental Results}


We construct three main tasks on MammalNet: 1) Standard animal and behavior classification on trimmed videos, 2) Compositional low-shot animal and behavior recognition on trimmed videos, and 3) Behavior detection on untrimmed videos. In both the classification and detection tasks, we baseline several state-of-the-art models~\cite{li2022mvitv2,feichtenhofer2019slowfast,actionformer} that have been successfully applied to human action recognition and detection. 
In the following, we describe the formulation of each challenge and provide baselines and analysis.

\subsection{Standard Animal and Behavior Classification on Trimmed Videos}
This task explores classification of both the primary behavior that occurs in a trimmed video and the animal that performs the behavior. We report the top-1 per-example and per-class accuracy for all the baseline models.

\noindent\textbf{Many, medium, few splits.} To capture the effect of the long-tailed nature of the MammalNet dataset, we group the animal, behavior, and their composition classes into \emph{many}, \emph{medium}, and \emph{few} based on their frequency, and report the average per-class accuracy bands chosen based on the frequency percentiles. For animal categories this is broken down as \emph{many}: top 7\% frequent classes, \emph{medium}: middle 16\% classes, and \emph{few}: the remaining 77\% classes. For behavior, \emph{many}: top 33\% frequent classes, \emph{medium}: middle 33\%, and \emph{few}: the remaining 33\% classes.  For joint classification, \emph{many}: top 3\% frequent classes, \emph{medium}: middle 17\% classes, \emph{few}: the remaining 80\% classes. We show the number of classes per each split in Table \ref{per_class_accuracy}. 


\noindent \textbf{Dataset setup.} 
We randomly split the examples from each animal-behavior category into 70\% for training, 10\% for validation, and 20\% for testing, and it results in 14,554 training, 1,638 validation, and 3,841 testing videos, respectively.


    

\begin{table*}[t!]
\begin{center}
\begin{tabular}{l|cccccc}
\toprule
& \multicolumn{6}{c}{Compositional Low-Shot Behavior Classification} \\
 \multirow{2}{*}{Baselines} & \multicolumn{2}{c}{0-shot} & \multicolumn{2}{c}{1-shot} & \multicolumn{2}{c}{5-shot} \\
    & Per-example  & Per-class & Per-example & Per-class & Per-example & Per-class \\
    &A / B & A / B & A / B & A / B & A / B & A / B \\
\midrule
  C3D* & 27.4 / 23.7& 18.3 / 16.1 & 29.2 / 25.5& 21.1 / 19.9 & 33.3 / 29.3 & 25.2 / 23.0 \\
  I3D* &  25.3 / 23.9 & 16.7 / 15.2 & 26.8 / 25.7 & 16.9 / 18.3  & 30.5 / 28.3 & 21.7 / 21.9\\
  SlowFast*  & 26.2 / 24.8 & 17.9 / 16.3 & 26.5 / 26.3 & 16.7 / 19.0 & 29.6 / 29.2& 22.5 / 22.4\\
  MViT V2*  & \textbf{32.2} / \textbf{26.2} & \textbf{20.7} / \textbf{18.1} & \textbf{33.7} / \textbf{28.9}& \textbf{22.9} / \textbf{22.5}& \textbf{39.3} / \textbf{31.7} & \textbf{31.0} / \textbf{26.0} \\
\bottomrule
\end{tabular}
\end{center}
\caption{Compositional low-shot animal and behavior recognition. * denotes the initialization from the model pretrained on Kinetics 400~\cite{kay2017kinetics}. ``A'' denotes the animal category and ``B'' denotes the behavior category. The best performance per each column has been highlighted in \textbf{bold}.} 
\label{low_shot_behavior}
\vspace{-1em}
\end{table*}

\begin{table}[t!]
\centering
\resizebox{1.0\linewidth}{!}{
\begin{tabular}{l|ccccccc}
\toprule
\multirow{2}{*}{Baselines}  & \multicolumn{6}{c}{mAP} \\
  &  0.50 & 0.60 & 0.70 & 0.80  & 0.90 & Avg. \\
 \midrule
  CoLA \cite{zhang2021cola}     & 26.02 & 22.70 &  18.98 & 13.46 & 3.05 &  15.81   \\
  TAGS \cite{tags}             & 23.09 & 20.97 & 19.09 & 16.98 &  \textbf{12.56}  & 17.63  \\
  ActionFormer \cite{actionformer}  &  \textbf{28.48} &  \textbf{26.14} &  \textbf{23.17} &  \textbf{18.69} & 10.48  &\textbf{20.07} \\
\bottomrule
\end{tabular}
}
\caption{The results for behavior detection. We report mAP at the IoU thresholds of [0.5:0.1:0.9]. Average mAP is computed by averaging different tIoU thresholds. } 
\label{behavior_detection}
\vspace{-1em}
\end{table}

\noindent \textbf{Baselines.} We compare SlowFast~\cite{feichtenhofer2019slowfast}, I3D~\cite{i3d}, C3D~\cite{c3d} and MViT V2~\cite{li2022mvitv2} models on our tasks. These models are evaluated in two versions: 1) Training with random initialization 2) Initializing with weights from a model pretrained on Kinetics 400~\cite{kay2017kinetics}.  These methods were originally designed for human action recognition and hence do not have an ability to predict both an action and a subject by default. To accommodate them into our joint prediction setting, we have two task heads, one for animal category recognition and one for behavior recognition. We compute the joint loss, $\mathcal{L}_{joint}$, for both animal and behavior classification as shown in Fig.~\ref{fig:teaser}. We tune all the hyper-parameters on the validation data. The final hyper-parameters for each model are provided in the supplement. The loss is defined as:
\begin{equation} \mathcal{L}_{joint} = -\frac{1}{M}\sum_i^M{y^a_{i} \text{log}(p^a_{i})}-\frac{1}{N}\sum_j^N{y^b_{j} \text{log}(p^b_{j})}
\label{joint_loss}
\end{equation}
where $M$ is the number of animal classes, and $N$ is the number of behavior classes, $p^a_i$ and $y^a_i$ denote the animal prediction probability and ground truth label for the category $i$, $p^b_y$ and $y^b_j$ denote the behavior prediction probability and ground truth label for the category $j$.

\noindent \textbf{Experimental results.} 
The results for per-class and per-example classification are summarized in Tables~\ref{per_class_accuracy} and~\ref{per-example-accuracy}, respectively. We find that MViT v2 is able to achieve competitive results for all the splits. The best top-1 joint per-class accuracy is 17.8 and per-example accuracy is 30.6, which points to significant room for improvement on these challenging tasks. We also observe that transfer learning from the model that is pretrained on Kinetics 400, a human action recognition dataset, improves both the animal and behavior classification accuracy (with MViT v2, this corresponds to a per-class accuracy gain from 19.7 to 32.5 for animal classification, and from 27.7 to 37.8 for behavior classification). Additionally, we find that performance gain from pretraining is higher for frequently occurring animals and behaviors, indicated by the results shown in $many$, $medium$, and $few$ splits.  Accurately predicting low-frequency animal and behavior categories remains a significant challenge.

\subsection{Compositional Low-shot Animal and Behavior Classification on Trimmed Videos}
It is hard to find a sufficient number of labeled behavior samples for all the animals in our collected taxonomy. For example, \emph{numbat} and \emph{florida panther} have very few behavior annotations. However, it is plausible to imagine that behaviors can be transferred among different animals that have similar appearances and movement styles. Also the animal recognition under different behaviors (hunting, fighting) can be mutually transferred. To investigate these phenomena, we design the compositional low-shot animal and behavior classification task.

\noindent \textbf{Dataset setup.} We first select the animal-behavior compositional classes that contain more than 5 examples and allocate 25\% of classes to the test set (4,088 videos). For the remaining classes (the other 75\% of classes and those classes with $\leq$ 5 examples), we randomly designate 90\% of classes as the training set and 10\% as the validation set (898 videos). Under the low-shot scheme, for each compositional class, we randomly sample 5 examples from the test set and move 0, 1, or 5 of them into the training set for the zero-shot, 1-shot, and 5-shot setup, respectively (14,377, 14,511 and 15,047 training videos).  The train, val and test sets consist of 983, 53, and 134 compositional classes, respectively. 

\noindent \textbf{Baselines.} We evaluate the compositional low-shot classification with the SlowFast \cite{feichtenhofer2019slowfast}, I3D \cite{i3d}, C3D \cite{c3d} and MViT V2 \cite{li2022mvitv2} models under the joint loss. We initialize these models with the weights pretrained on Kinetics 400~\cite{kay2017kinetics}.

\noindent \textbf{Experimental results.} We summarize the results in Table~\ref{low_shot_behavior}. It shows that MViT v2 consistently achieves the best performance under our low-shot setup. The behavior classification can still achieve 26.2 per-example  and 18.1 top-1 accuracy under the zero-shot setting for MViT v2 model. This indicates that behavior classification is transferable from other animals in some cases. Additionally, we observe that its performance is improved to 31.7 per-example and 26.0 per-class top-1 accuracy under 5-shot setup, indicating that training on more videos with the behaviors from the same animal is still necessary. A similar phenomenon is observed to the few-shot animal recognition.

\begin{figure*}[t!]
\centering
\includegraphics[width=0.9\linewidth]
                  {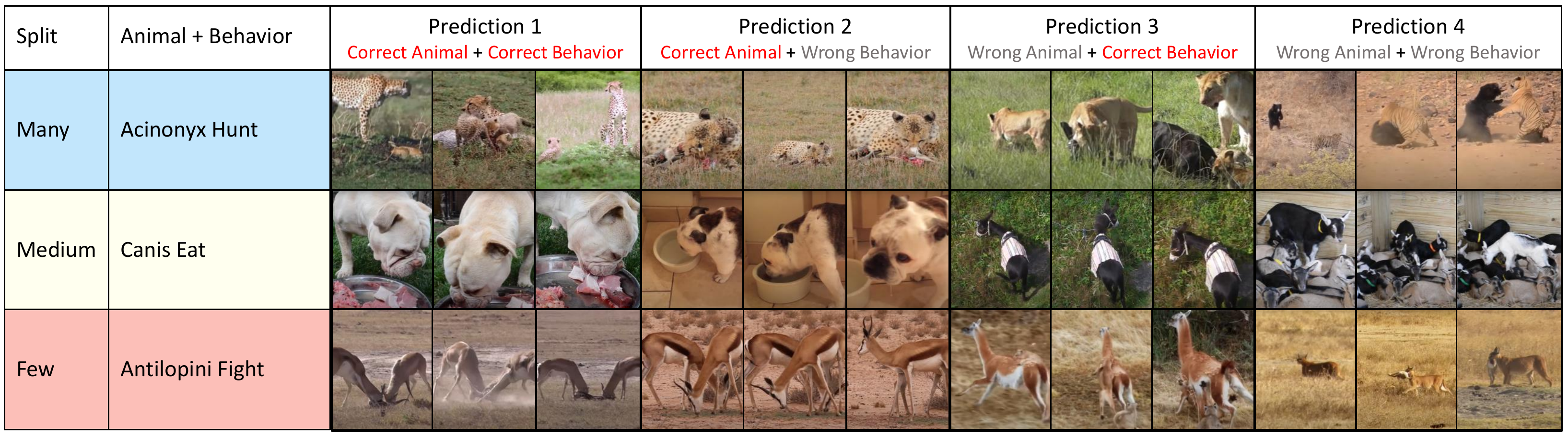}
\caption{The visualization presents various instances of joint animal and behavior classification. In the third column, accurate predictions are displayed, while the fourth, fifth, and sixth columns showcase mispredicted examples where either the animal or the behavior does not correspond to the correct prediction.
}
\label{vis_classification}
\vspace{-1.5em}
\end{figure*}
\subsection{Behavior Detection on Untrimmed Videos}
This task is to detect the behavior in untrimmed videos. The behavior detection algorithm should correctly detect the temporal range for the primary activity presented in the video. We follow previous temporal action localization works~\cite{caba2015activitynet,actionformer} to benchmark this task.  We report the mean Average Precision (mAP) with different temporal intersections over the union (tIoU) thresholds [0.5:0.1:0.9] as our evaluation metric. We also report the average mAP averaging across different tIoUs. 

\begin{figure}[t!]
\centering
\includegraphics[width=1.0\linewidth]
                  {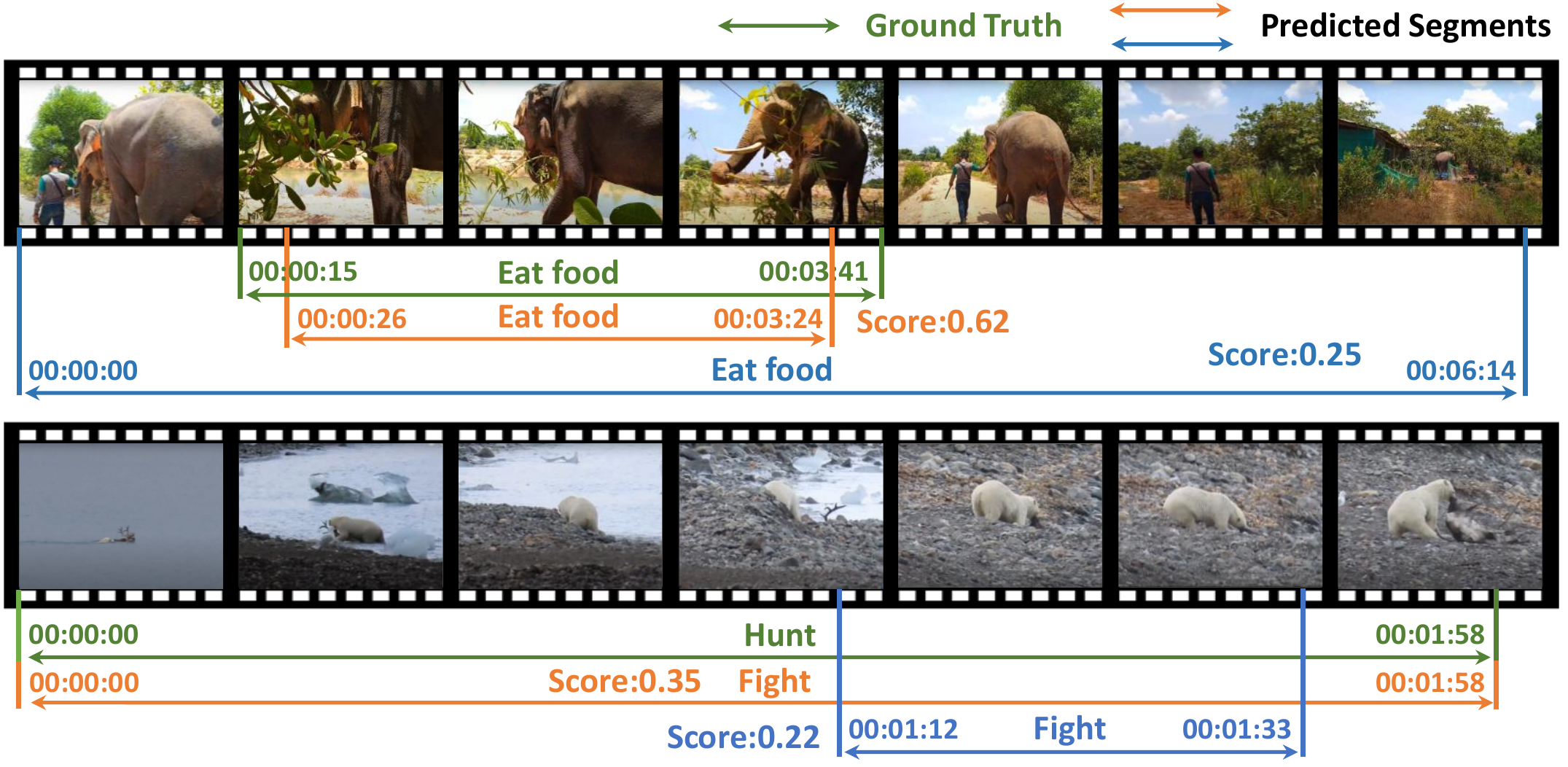}
\caption{The visualization for behavior detection examples.}
\label{vis_detection}
\vspace{-1.5em}
\end{figure}


\noindent \textbf{Dataset setup.} Similar to the previous standard animal and behavior classification, we follow the [train:0.7, val:0.1, test:0.2] ratios to split the untrimmed videos at the animal-behavior composition level. This results in 13,318 videos for training, 1,486 for validation and 3,542 for testing. We tune all the hyper-parameters based on the validation set.

\noindent \textbf{Baselines.} We evaluate our datasets with the baseline models such as ActionFormer~\cite{actionformer}, TAGS~\cite{tags}, and CoLA~\cite{zhang2021cola}. To produce the features for our MammalNet videos, we first finetune a two-stream I3D~\cite{i3d} model, that is originally pretrained on ImageNet~\cite{deng2009imagenet} and Kinetics 400~\cite{kay2017kinetics}, on our dataset, and then extract the RGB and optical flow features for each video. We concatenate these two features together as the model input. 

\noindent \textbf{Experimental results.} We show the behavior detection results in Table~\ref{behavior_detection}. Among all the baselines, ActionFormer demonstrates the most competitive performance with an average mAP of 20.07, and also achieves 28.48 mAP for the threshold of 0.5. Overall, it is clear to see that the behavior detection task is still very challenging for current methods.

\section{Analysis}

\noindent \textbf{Demonstration of animal and behavior classification.} We sample some prediction examples from MViT v2~\cite{li2022mvitv2} for the joint categories \emph{Acinonyx hunt}, \emph{Canis eat} and \emph{Antilopini fight}. We demonstrate one correct prediction and three mispredicted examples where the model mis-recognizes the animal or behavior or both of them in Fig.~\ref{vis_classification}. 


\noindent \textbf{Demonstration of behavior detection:} We visualize  behavior detection results in Fig.~\ref{vis_detection} from ActionFormer~\cite{actionformer}. The top example shows correctly predicted behavior with a proposal closely aligned with the ground-truth. The bottom example shows a misclassified behavior, and it mistakenly predicts the \emph{hunt} behavior as \emph{fight}.

\noindent \textbf{Joint animal and behavior recognition vs. separate recognition.} 
We also train the model for recognizing the animal and behavior separately and provide the results in the Table 1 of the supplementary. Comparing with the joint recognition, we find that training a system to recognize the animal and behavior together can improve the behavior recognition under separate training by $\sim$2.2 per-class accuracy. This indicates that being capable of understanding the animal types can benefit behavior prediction. However, the results also indicate that predicting animal only can be more useful in recognizing animal types.

\noindent \textbf{Dataset bias.} Our dataset is downloaded from YouTube channels with diverse backgrounds and video quality, and it might differ from video datasets collected for behavioral or ecological studies. The videos on YouTube might also exhibit potential biases: 

\noindent 1) Over-representation of some behaviors and under-representation of others. For example, people might prefer watching videos with \emph{fight} or \emph{eat food} activities, and these two behaviors are more likely to be over-represented, while \emph{urinate} and \emph{defecate} behaviors are less interesting to humans and hence are underrepresented on YouTube. However, they are still important in informing conservation-related actions to protect the environment.

\noindent 2) Bias towards captive animals or wild animals that are habituated to humans. We find that many videos are shot at zoos, farms, and homes, etc. These animals may display different behaviors than wild or non-habituated animals.

\section{Conclusion}

We introduced MammalNet, a large-scale  video dataset for mammal recognition and behavior understanding. We have collected videos for hundreds of different mammals and structured them by following the scientific mammal taxonomy. MammalNet consists of 18,346 untrimmed videos covering 173 mammal categories and 12 common behaviors. 
We established three challenges: standard animal and behavior classification, compositional low-shot animal and behavior classification, and behavior detection. 
Through our experiments, we found that the accurate recognition of animals at scale and their common behaviors is very challenging even with current state-of-the-art models, especially when the dataset has a long-tail distribution. We also found that learning to recognize the behavior of unseen animals is possible via transfer from the other seen animals. To promote further research and development in the field of animal behavior study, we have open-sourced all of our data and code to the research community.\newline

\noindent \textbf{Acknowledgement}: This work is supported by  KAUST BAS/1/1685-01-01 and KAUST FCC/1/1973-58-01 (Red Sea Research Center), DARPA’s SemaFor and PTG programs, the Caltech Resnick Sustainability Institute, and Germany's Excellence Strategy–`Centre for the Advanced Study of Collective Behaviour' EXC 2117-422037984.

{\small
\bibliographystyle{ieee_fullname}
\bibliography{egbib}

\begin{thebibliography}{10}\itemsep=-1pt

\bibitem{animal_a_z}
A-z animals.
\newblock \url{https://a-z-animals.com/}.

\bibitem{national_geographic}
National geographics.
\newblock \url{https://www.nationalgeographic.com/}.

\bibitem{alcock2009animal}
John Alcock.
\newblock {\em Animal behavior: An evolutionary approach}.
\newblock Sinauer Associates, 2009.

\bibitem{anderson2014toward}
David~J Anderson and Pietro Perona.
\newblock Toward a science of computational ethology.
\newblock {\em Neuron}, 84(1):18--31, 2014.

\bibitem{bala2020automated}
Praneet~C Bala, Benjamin~R Eisenreich, Seng Bum~Michael Yoo, Benjamin~Y Hayden,
  Hyun~Soo Park, and Jan Zimmermann.
\newblock Automated markerless pose estimation in freely moving macaques with
  openmonkeystudio.
\newblock {\em Nature communications}, 11(1):1--12, 2020.

\bibitem{animal_behavior}
Oded Berger-Tal, Daniel~T Blumstein, Scott Carroll, Robert~N Fisher, Sarah~L
  Mesnick, Megan~A Owen, David Saltz, Colleen~Cassady St~Claire, and Ronald~R
  Swaisgood.
\newblock A systematic survey of the integration of animal behavior into
  conservation.
\newblock {\em Conservation Biology}, 30(4):744--753, 2016.

\bibitem{beyan2013detection}
Cigdem Beyan and Robert~B Fisher.
\newblock Detection of abnormal fish trajectories using a clustering based
  hierarchical classifier.
\newblock In {\em BMVC}, 2013.

\bibitem{breed2021animal}
Michael~D Breed and Janice Moore.
\newblock {\em Animal behavior}.
\newblock Academic Press, 2021.

\bibitem{caba2015activitynet}
Fabian Caba~Heilbron, Victor Escorcia, Bernard Ghanem, and Juan Carlos~Niebles.
\newblock Activitynet: A large-scale video benchmark for human activity
  understanding.
\newblock In {\em Proceedings of the ieee conference on computer vision and
  pattern recognition}, pages 961--970, 2015.

\bibitem{cao2019cross}
Jinkun Cao, Hongyang Tang, Hao-Shu Fang, Xiaoyong Shen, Cewu Lu, and Yu-Wing
  Tai.
\newblock Cross-domain adaptation for animal pose estimation.
\newblock In {\em Proceedings of the IEEE/CVF International Conference on
  Computer Vision}, pages 9498--9507, 2019.

\bibitem{i3d}
Joao Carreira and Andrew Zisserman.
\newblock Quo vadis, action recognition? a new model and the kinetics dataset.
\newblock In {\em proceedings of the IEEE Conference on Computer Vision and
  Pattern Recognition}, pages 6299--6308, 2017.

\bibitem{ceballos2020vertebrates}
Gerardo Ceballos, Paul~R Ehrlich, and Peter~H Raven.
\newblock Vertebrates on the brink as indicators of biological annihilation and
  the sixth mass extinction.
\newblock {\em Proceedings of the National Academy of Sciences},
  117(24):13596--13602, 2020.

\bibitem{chen2016locality}
Yu-Chen Chen, Shintami~C Hidayati, Wen-Huang Cheng, Min-Chun Hu, and Kai-Lung
  Hua.
\newblock Locality constrained sparse representation for cat recognition.
\newblock In {\em International Conference on Multimedia Modeling}, pages
  140--151. Springer, 2016.

\bibitem{defays1977efficient}
Daniel Defays.
\newblock An efficient algorithm for a complete link method.
\newblock {\em The Computer Journal}, 20(4):364--366, 1977.

\bibitem{deng2009imagenet}
Jia Deng, Wei Dong, Richard Socher, Li-Jia Li, Kai Li, and Li Fei-Fei.
\newblock Imagenet: A large-scale hierarchical image database.
\newblock In {\em 2009 IEEE conference on computer vision and pattern
  recognition}, pages 248--255. Ieee, 2009.

\bibitem{dishman2014self}
Rod~K Dishman, Andrew~S Jackson, and Molly~S Bray.
\newblock Self-regulation of exercise behavior in the tiger study.
\newblock {\em Annals of Behavioral Medicine}, 48(1):80--91, 2014.

\bibitem{dugatkin2020principles}
Lee~Alan Dugatkin.
\newblock {\em Principles of animal behavior}.
\newblock University of Chicago Press, 2020.

\bibitem{fang2021pose}
Cheng Fang, Tiemin Zhang, Haikun Zheng, Junduan Huang, and Kaixuan Cuan.
\newblock Pose estimation and behavior classification of broiler chickens based
  on deep neural networks.
\newblock {\em Computers and Electronics in Agriculture}, 180:105863, 2021.

\bibitem{feichtenhofer2019slowfast}
Christoph Feichtenhofer, Haoqi Fan, Jitendra Malik, and Kaiming He.
\newblock Slowfast networks for video recognition.
\newblock In {\em Proceedings of the IEEE/CVF international conference on
  computer vision}, pages 6202--6211, 2019.

\bibitem{feng2021action}
Liqi Feng, Yaqin Zhao, Yichao Sun, Wenxuan Zhao, and Jiaxi Tang.
\newblock Action recognition using a spatial-temporal network for wild felines.
\newblock {\em Animals}, 11(2):485, 2021.

\bibitem{graving2019deepposekit}
Jacob~M Graving, Daniel Chae, Hemal Naik, Liang Li, Benjamin Koger, Blair~R
  Costelloe, and Iain~D Couzin.
\newblock Deepposekit, a software toolkit for fast and robust animal pose
  estimation using deep learning.
\newblock {\em Elife}, 8:e47994, 2019.

\bibitem{gupta2021dftnet}
Shilpi Gupta, Prerana Mukherjee, Santanu Chaudhury, Brejesh Lall, and Hemanth
  Sanisetty.
\newblock Dftnet: Deep fish tracker with attention mechanism in unconstrained
  marine environments.
\newblock {\em IEEE Transactions on Instrumentation and Measurement}, 70:1--13,
  2021.

\bibitem{heilbron2014collecting}
Fabian~Caba Heilbron and Juan~Carlos Niebles.
\newblock Collecting and annotating human activities in web videos.
\newblock In {\em Proceedings of International Conference on Multimedia
  Retrieval}, pages 377--384, 2014.

\bibitem{kay2017kinetics}
Will Kay, Joao Carreira, Karen Simonyan, Brian Zhang, Chloe Hillier, Sudheendra
  Vijayanarasimhan, Fabio Viola, Tim Green, Trevor Back, Paul Natsev, et~al.
\newblock The kinetics human action video dataset.
\newblock {\em arXiv preprint arXiv:1705.06950}, 2017.

\bibitem{kays2022mammals}
Roland Kays, Monica Lasky, Maximilian~L Allen, Robert~C Dowler, Melissa~TR
  Hawkins, Andrew~G Hope, Brooks~A Kohli, Verity~L Mathis, Bryan McLean, Link~E
  Olson, et~al.
\newblock Which mammals can be identified from camera traps and crowdsourced
  photographs?
\newblock {\em Journal of Mammalogy}, 2022.

\bibitem{khan2020animalweb}
Muhammad~Haris Khan, John McDonagh, Salman Khan, Muhammad Shahabuddin, Aditya
  Arora, Fahad~Shahbaz Khan, Ling Shao, and Georgios Tzimiropoulos.
\newblock Animalweb: A large-scale hierarchical dataset of annotated animal
  faces.
\newblock In {\em Proceedings of the IEEE/CVF Conference on Computer Vision and
  Pattern Recognition}, pages 6939--6948, 2020.

\bibitem{khosla2011novel}
Aditya Khosla, Nityananda Jayadevaprakash, Bangpeng Yao, and Fei-Fei Li.
\newblock Novel dataset for fine-grained image categorization: Stanford dogs.
\newblock In {\em Proc. CVPR workshop on fine-grained visual categorization
  (FGVC)}, volume~2. Citeseer, 2011.

\bibitem{laws2007case}
Nicole Laws, Andre Ganswindt, Michael Heistermann, Moira Harris, Stephen
  Harris, and Chris Sherwin.
\newblock A case study: fecal corticosteroid and behavior as indicators of
  welfare during relocation of an asian elephant.
\newblock {\em Journal of Applied Animal Welfare Science}, 10(4):349--358,
  2007.

\bibitem{li2020wildlife}
Weining Li, Sirnam Swetha, and Mubarak Shah.
\newblock Wildlife action recognition using deep learning.
\newblock 2020.

\bibitem{li2022mvitv2}
Yanghao Li, Chao-Yuan Wu, Haoqi Fan, Karttikeya Mangalam, Bo Xiong, Jitendra
  Malik, and Christoph Feichtenhofer.
\newblock Mvitv2: Improved multiscale vision transformers for classification
  and detection.
\newblock In {\em Proceedings of the IEEE/CVF Conference on Computer Vision and
  Pattern Recognition}, pages 4804--4814, 2022.

\bibitem{mathis2021pretraining}
Alexander Mathis, Thomas Biasi, Steffen Schneider, Mert Yuksekgonul, Byron
  Rogers, Matthias Bethge, and Mackenzie~W Mathis.
\newblock Pretraining boosts out-of-domain robustness for pose estimation.
\newblock In {\em Proceedings of the IEEE/CVF Winter Conference on Applications
  of Computer Vision}, pages 1859--1868, 2021.

\bibitem{mathis2020deep}
Mackenzie~Weygandt Mathis and Alexander Mathis.
\newblock Deep learning tools for the measurement of animal behavior in
  neuroscience.
\newblock {\em Current opinion in neurobiology}, 60:1--11, 2020.

\bibitem{mench1998important}
Joy Mench.
\newblock Why it is important to understand animal behavior.
\newblock {\em ILAR journal}, 39(1):20--26, 1998.

\bibitem{mu2020learning}
Jiteng Mu, Weichao Qiu, Gregory~D Hager, and Alan~L Yuille.
\newblock Learning from synthetic animals.
\newblock In {\em Proceedings of the IEEE/CVF Conference on Computer Vision and
  Pattern Recognition}, pages 12386--12395, 2020.

\bibitem{tags}
Sauradip Nag, Xiatian Zhu, Yi-Zhe Song, and Tao Xiang.
\newblock Temporal action detection with global segmentation mask learning.
\newblock {\em European Conference on Computer Vision}, 2022.

\bibitem{ng2022animal}
Xun~Long Ng, Kian~Eng Ong, Qichen Zheng, Yun Ni, Si~Yong Yeo, and Jun Liu.
\newblock Animal kingdom: A large and diverse dataset for animal behavior
  understanding.
\newblock In {\em Proceedings of the IEEE/CVF Conference on Computer Vision and
  Pattern Recognition}, pages 19023--19034, 2022.

\bibitem{owoeye2018online}
Kehinde Owoeye and Stephen Hailes.
\newblock Online collective animal movement activity recognition.
\newblock {\em arXiv preprint arXiv:1811.09067}, 2018.

\bibitem{rahman2014fast}
Shah~Atiqur Rahman, Insu Song, Maylor~KH Leung, Ickjai Lee, and Kyungmi Lee.
\newblock Fast action recognition using negative space features.
\newblock {\em Expert Systems with Applications}, 41(2):574--587, 2014.

\bibitem{rashid2017interspecies}
Maheen Rashid, Xiuye Gu, and Yong Jae~Lee.
\newblock Interspecies knowledge transfer for facial keypoint detection.
\newblock In {\em Proceedings of the IEEE Conference on Computer Vision and
  Pattern Recognition}, pages 6894--6903, 2017.

\bibitem{robinson2014comparison}
Lianne Robinson and Gernot Riedel.
\newblock Comparison of automated home-cage monitoring systems: emphasis on
  feeding behaviour, activity and spatial learning following pharmacological
  interventions.
\newblock {\em Journal of neuroscience methods}, 234:13--25, 2014.

\bibitem{shooter2021sydog}
Moira Shooter, Charles Malleson, and Adrian Hilton.
\newblock Sydog: A synthetic dog dataset for improved 2d pose estimation.
\newblock {\em arXiv preprint arXiv:2108.00249}, 2021.

\bibitem{c3d}
Du Tran, Lubomir Bourdev, Rob Fergus, Lorenzo Torresani, and Manohar Paluri.
\newblock Learning spatiotemporal features with 3d convolutional networks.
\newblock In {\em Proceedings of the IEEE international conference on computer
  vision}, pages 4489--4497, 2015.

\bibitem{tuia2022perspectives}
Devis Tuia, Benjamin Kellenberger, Sara Beery, Blair~R Costelloe, Silvia Zuffi,
  Benjamin Risse, Alexander Mathis, Mackenzie~W Mathis, Frank van Langevelde,
  Tilo Burghardt, et~al.
\newblock Perspectives in machine learning for wildlife conservation.
\newblock {\em Nature communications}, 13(1):1--15, 2022.

\bibitem{van2015building}
Grant Van~Horn, Steve Branson, Ryan Farrell, Scott Haber, Jessie Barry, Panos
  Ipeirotis, Pietro Perona, and Serge Belongie.
\newblock Building a bird recognition app and large scale dataset with citizen
  scientists: The fine print in fine-grained dataset collection.
\newblock In {\em Proceedings of the IEEE Conference on Computer Vision and
  Pattern Recognition}, pages 595--604, 2015.

\bibitem{inaturalist}
Grant Van~Horn, Oisin Mac~Aodha, Yang Song, Yin Cui, Chen Sun, Alex Shepard,
  Hartwig Adam, Pietro Perona, and Serge Belongie.
\newblock The inaturalist species classification and detection dataset.
\newblock In {\em Proceedings of the IEEE conference on computer vision and
  pattern recognition}, pages 8769--8778, 2018.

\bibitem{vanexploring}
Grant Van~Horn, Rui Qian, Kimberly Wilber, Hartwig Adam, Oisin Mac~Aodha, and
  Serge Belongie.
\newblock Exploring fine-grained audiovisual categorization with the ssw60
  dataset.

\bibitem{van2022ssw60}
Grant Van~Horn, Rui Qian, Kimberly Wilber, Hartwig Adam, Oisin Mac~Aodha, and
  Serge Belongie.
\newblock Exploring fine-grained audiovisual categorization with the ssw60
  dataset, 2022.

\bibitem{von2021big}
Lukas von Ziegler, Oliver Sturman, and Johannes Bohacek.
\newblock Big behavior: challenges and opportunities in a new era of deep
  behavior profiling.
\newblock {\em Neuropsychopharmacology}, 46(1):33--44, 2021.

\bibitem{vondrick2013efficiently}
Carl Vondrick, Donald Patterson, and Deva Ramanan.
\newblock Efficiently scaling up crowdsourced video annotation.
\newblock {\em International journal of computer vision}, 101(1):184--204,
  2013.

\bibitem{welinder2010caltech}
Peter Welinder, Steve Branson, Takeshi Mita, Catherine Wah, Florian Schroff,
  Serge Belongie, and Pietro Perona.
\newblock Caltech-ucsd birds 200.
\newblock 2010.

\bibitem{yang2016human}
Heng Yang, Renqiao Zhang, and Peter Robinson.
\newblock Human and sheep facial landmarks localisation by triplet interpolated
  features.
\newblock In {\em 2016 IEEE Winter Conference on Applications of Computer
  Vision (WACV)}, pages 1--8. IEEE, 2016.

\bibitem{zhang2021cola}
Can Zhang, Meng Cao, Dongming Yang, Jie Chen, and Yuexian Zou.
\newblock Cola: Weakly-supervised temporal action localization with snippet
  contrastive learning.
\newblock In {\em Proceedings of the IEEE/CVF Conference on Computer Vision and
  Pattern Recognition (CVPR)}, pages 16010--16019, June 2021.

\bibitem{actionformer}
Chenlin Zhang, Jianxin Wu, and Yin Li.
\newblock Actionformer: Localizing moments of actions with transformers.
\newblock In {\em European Conference on Computer Vision}, 2022.

\end{thebibliography}
}

\end{document}